\documentclass[11pt]{article}
\usepackage{coling2020}
\usepackage{times}
\usepackage{url}
\usepackage{hyperref}
\usepackage{latexsym}
\usepackage{color, colortbl}
\definecolor{Gray}{gray}{0.9}
\usepackage{array}
\newcolumntype{R}[1]{>{\raggedleft\let\newline\\\arraybackslash\hspace{0pt}}m{#1}}
\newcolumntype{L}[1]{>{\raggedright\let\newline\\\arraybackslash\hspace{0pt}}m{#1}}
\usepackage{dirtytalk}
\usepackage{listings}
\usepackage{graphicx}
\usepackage{placeins}
\usepackage{multicol}
\usepackage{pgfplots}
\usepackage{booktabs}
\usepackage{caption}
\usepackage{subcaption}

\pgfplotsset{compat=1.9}

\definecolor{red}{HTML}{ab0000}
\definecolor{grey}{HTML}{ababab}
\definecolor{green}{HTML}{00ab00}
\definecolor{darkblue}{HTML}{0000ab}
\definecolor{lightblue}{HTML}{69abff}
\definecolor{darkorange}{HTML}{d47e00}

\colingfinalcopy

\title{FiSSA at SemEval-2020 Task 9: Fine-tuned For Feelings}

\usepackage{enumitem}

\setlist[itemize]{leftmargin=*}

\author{Bertelt Braaksma \hspace{0.4em} Richard Scholtens \hspace{0.4em} Stan van Suijlekom \hspace{0.4em} Remy Wang \hspace{0.4em} Ahmet \"{U}st\"{u}n
\vspace{.2cm}
\\
University of Groningen \\
Groningen, the Netherlands \\
\hspace{-15px}
{\tt \fontsize{8,6}{6} \selectfont \{b.braaksma.1, j.f.p.scholtens, s.van.suijlekom, r.wang.7\}@student.rug.nl, a.ustun@rug.nl}
\\}

\date{}

\begin{document}
\maketitle


\begin{abstract}


In this paper, we present our approach for sentiment classification on Spanish-English code-mixed social media data in the SemEval-2020 Task 9. We investigate performance of various pre-trained Transformer models by using different fine-tuning strategies. We explore both monolingual and multilingual models with the standard fine-tuning method. Additionally, we propose a custom model that we fine-tune in two steps: once with a language modeling objective, and once with a task-specific objective. Although two-step fine-tuning improves sentiment classification performance over the base model, the large multilingual XLM-RoBERTa model achieves best weighted F1-score with 0.537 on development data and 0.739 on test data. With this score, our team \textit{jupitter} placed tenth overall in the competition.




\end{abstract}

\section{Introduction}
\blfootnote{\hspace{-17px}This work is licensed under a Creative Commons Attribution 4.0 International License.\\License details: \url{http://creativecommons.org/licenses/by/4.0/}.}



Code-mixing is a phenomenon in which two or more languages are used in a single utterance. It occurs at various levels of linguistic structure: across sentences (i.e., inter-sentential), within a sentence (i.e.,intra-sentential), or at the word/morpheme level. In addition to spoken language, this phenomenon has become especially prevalent on social media. As monolingual systems cannot deal with the code-mixed data, it poses a major challenge for even the most standard NLP tasks. To this end, SemEval 2020 Task 9 \cite{patwa2020sentimix} proposes the sentiment analysis task for code-mixed social media text, specifically on English-Spanish (Spanglish) and English-Hindi (Hinglish) language pairs. 


In this paper, we present our approach called Fine-tuned Spanglish Sentiment Analysis, or FiSSA for short. We focus on various pre-trained language models for Spanglish sentiment classification by fine-tuning their contextualized word embeddings. By doing so, we examine two challenging aspects of code-mixed language processing: (a) Multilinguality, (b) Domain. We firstly compare monolingual models with their multilingual counterparts to evaluate the multilingual solution on code-switching data, considering the first aspect. Secondly, to see the domain effect, we further fine-tune the multilingual model on domain-specific unlabeled data. Finally, we use the most recent state-of-the-art pre-trained model and compare it to our custom model with domain information. 


Our research shows that fine-tuning a pre-trained language model is a good choice compared to the standard BLSTM model when training data is limited. However, on code-mixed data, monolingual pre-trained models tend to perform better on different portions of the data depending on the use of languages. As the best alternative, a large multilingual model provides better generalization and results in a stronger performance. 



\section{Background}
\paragraph{Code-Mixed Text Processing}

Only a limited amount of research has been done in the field of sentiment analysis on Spanish-English social media data. However, some writing has been done regarding Spanish-English code-mixing for other NLP tasks, such as part-of-speech tagging \cite{solorio2008} and language identification \cite{solorio-etal-2014-overview}. In the first shared task of language identification on code-switched data including Spanish-English (SPA-EN), many systems benefited from a combination of machine learning methods such as an SVM, a CRF, an extended Markov Model, and hand crafted or frequency-based features. On SPA-EN, the best performing system employed a CRF classifier by using various character- and word-level features together with external resources that include monolingual corpora with named-entity lists \cite{bar-dershowitz-2014-tel}. 

With regard to English-Spanish sentiment analysis on social media data, the first English-Spanish code-switching Twitter corpus annotated with sentiment labels was made available in the research conducted by \newcite{vilares-etal-2015-sentiment} and \newcite{vilares-etal-2016-en}. In their trinary annotated corpus, a collection of 3062 tweets were annotated by three annotators fluent in both English and Spanish, classifying each tweet as either \textit{positive}, \textit{neutral} or \textit{negative}. They discovered that there was a small advantage to be gained from using a multilingual approach. However, both monolingual and multilingual approaches struggled with code-switching text.

\paragraph{Pre-trained Transformers}

Deep, Transformer \cite{DBLP:journals/corr/VaswaniSPUJGKP17} based language models provide general-purpose contextualized linguistics representations that have shown great success on various NLP tasks \cite{devlin2018bert,yang2019xlnet,liu2019roberta}. These models are pre-trained on large unannotated corpora, and then fine-tuned for downstream tasks according to task-specific objectives. As well as monolingual Transformer models, multilingual models that are pre-trained on the concatenation of monolingual corpora from multiple languages, have enabled significant advances in multilingual NLP \cite{devlin2018bert,lample2019cross,conneau2019unsupervised}. \newcite{pires-etal-2019-multilingual} showed that multilingual BERT \cite{devlin2018bert} provides a strong cross-lingual generalization, which allows for the incorporation of information from multiple languages, for example in a code-switching scenario. 



\section{System overview}

\subsection{Baseline}
As a baseline, we used a standard bidirectional LSTM \cite{doi:10.1162/neco.1997.9.8.1735} with pre-trained word embeddings. In order to combine English and Spanish words in the BLSTM, we use stacked embeddings which were a mix of Flair English and Flair Spanish word embeddings \cite{akbik2019naacl}. 


\subsection{Pre-trained Transformer Models}

For our main system, we incorporated different pre-trained language models by fine-tuning them for the sentiment analysis. We picked a selection of monolingual and multilingual models to see how they perform differently on code-switched data. For our monolingual models, we used English BERT-base \cite{devlin2018bert} and Spanish BERT-base \cite{CaneteCFP2020}. They both have the same `base' architecture, consisting of 12 Transformer blocks with
12 self-attention heads and hidden size of 768. The English model has a 30k WordPiece (WP) vocabulary \cite{wu2016google} and the Spanish model has a 31k SentencePiece (SP) vocabulary \cite{kudo-richardson-2018-sentencepiece}.

For our multilingual models we used multilingual BERT (M-BERT) with a 110k shared WP vocabulary and XLM-RoBERTa (XLM-R) large with a 250k SP vocabulary \cite{liu2019roberta}. Both models were trained on a concatenation of over 100 languages. However, M-BERT's architecture is the same as that of the monolingual `base' models, whilst XLM-R has a larger network with 24 Transformer blocks of 16 self-attention heads and hidden size of 1024.

Besides the off-the-shelf models, we also provided a domain-specific custom model by fine-tuning M-BERT with a language modeling objective on the training data. 
For the task-specific fine-tuning, we applied a softmax classifier over the pooled output the of first token ([CLS]), which gives the sentence representation.



\section{Experimental setup}

\subsection{Data}
\begin{figure}[t]
    \centering
    \begin{subfigure}[b]{0.4\textwidth}
        \begin{tikzpicture}
        \pgfplotsset{width=198pt}
        \centering
            \begin{axis}[
                ybar,
                /pgf/bar width=15pt,
                enlarge x limits=0.45,
                symbolic x coords={train, dev},
                xtick=data,
                tick pos=left,
                ymin=0,
                ymax=7500,
                ]
            
                \addplot[red!80!white,fill=red!80!white]
                    	coordinates {(train, 2023)
                    	             (dev, 0)};
                \addplot[orange!80!white,fill=orange!80!white]
                    	coordinates {(train, 3794)
                    	             (dev, 0)};
                \addplot[green!80!white,fill=green!80!white]
                    	coordinates {(train, 6005)
                    	             (dev, 0)};
            
            \end{axis}
            \begin{axis}[
                ybar,
                /pgf/bar width=15pt,
                enlarge x limits=0.45,
                legend style={at={(0.5, 1.003)},
                  anchor=north,legend columns=-1},
                axis y line*=right,
                symbolic x coords={train, dev},
                xtick=data,
                ymin=0,
                ymax=1900,
                xticklabels={,,},
                xtick pos=left,
                ]
                \addplot[red!80!white,fill=red!80!white]
                    	coordinates {(train, 0)
                    	             (dev, 506)};
                \addplot[orange!80!white,fill=orange!80!white]
                    	coordinates {(train, 0)
                    	             (dev, 994)};
                \addplot[green!80!white,fill=green!80!white]
                    	coordinates {(train, 0)
                    	             (dev, 1498)};
                \legend{Negative, Neutral, Positive}
            \end{axis}
        \end{tikzpicture}
        \caption{Distribution of sentiment labels in both datasets}
        \label{fig:distribution}
    \end{subfigure}
    \begin{subfigure}[b]{0.5\textwidth}
        \centering
        \begin{tabular}{L{2.5cm}r}\toprule
        Label       & Count \\\midrule
        lang1       & 7,693  \\ 
        lang2       & 25,435 \\ 
        ambiguous   & 43    \\
        other       & 11,691 \\
        mixed       & 19    \\
        ne          & 1,232  \\
        unk         & 47    \\
        fw          & 4     \\\bottomrule
        \end{tabular}
                \vspace{0.35cm} 
        \caption{Distribution of token-level labels}
                \vspace{0.2cm} 
        \label{tab:count}
    \end{subfigure}
    \caption{Details of the training and the development datasets}
\end{figure}

We used the training and development datasets provided by the shared-task \cite{patwa2020sentimix}. For comparison, we split the training data into two pieces: 90\% for training and 10\% for development. The development data provided by the organizers was then used for the evaluation of all of our models, as the test set was not released. For our final submission, we trained the models by using the whole training set, and used the development data as such.

The three sentiment labels were not equally distributed in the training and development datasets. Positive tweets were over-represented, with roughly 6,000 tweets in training and over 1,400 in development. The second biggest class was neutral with almost 4,000 and 1,000 tweets in training and development respectively. Negative tweets formed the smallest class, with roughly 2,000 tweets in training, and over 500 in development. However, Figure~\ref{fig:distribution} shows that the distribution was similarly skewed in both datasets.

Both datasets were provided with word-level labels including language ids. Figure \ref{tab:count} shows the label distribution in the datasets. 


\subsection{Implementation}
The baseline system was developed using the Flair library\footnote{flairNLP, version 0.4.5, URL: \url{https://github.com/flairNLP/flair}}. For our LSTM, we used the word embeddings included in the library. In our case, the English and Spanish forward and backward embeddings were used, which were trained on a billion word corpus and Wikipedia respectively. The Flair LSTM classifier was trained using a learning rate of 0.1 and a mini batch size of 32. 

For fine-tuning the pre-trained Transformer models, we used the HuggingFace Transformers\footnote{\hspace{1pt}\includegraphics[height=6.5pt]{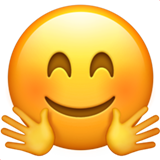} Transformers, version 2.5.0, URL: \url{https://huggingface.co/transformers/v2.5.0/}} library \cite{Wolf2019HuggingFacesTS}. The existing code had to be slightly modified to add support for sentiment analysis. We fine-tuned all of the models with the exact same hyper-parameters. We set the Adam epsilon to $ 1^{-8}$ and learning rate to $ 1^{- 5}$, and fine-tuned for 3 epochs\footnote{The code for our best performing model can be found at \url{https://github.com/barfsma/FiSSA}.}.

\section{Results}



The results of each model are presented in Table~\ref{tab:hf_all_results}. Precision, recall, and F1-scores are all weighted scores. As the table shows, all of our fine-tuned models performed better than the baseline BLSTM system in every regard, and XLM-RoBERTa large was the best overall, with the highest weighted F1-score. Interestingly, when we shift our focus to the monolingual models, we see that the English model performed worse than the Spanish model. The multilingual BERT-model sits right in the middle of these two when we look at the performance metrics. 

\begin{table}[h]
    \centering
    \begin{tabular}{R{0.25cm}lcccc}\toprule
      & Model                                   & Precision & Recall & F1-score & Accuracy \\\midrule
    1 & BLSTM                                   & 0.395 & 0.355 & 0.325 & 0.395 \\
    2 & English BERT-base                       & 0.503 & 0.515 & 0.506 & 0.515 \\
    3 & Spanish BERT-base                       & 0.514 & 0.526 & 0.517 & 0.526 \\
    4 & Multilingual BERT-base                  & 0.506 & 0.519 & 0.510 & 0.519 \\
    5 & Custom BERT-base                        & 0.513 & 0.526 & 0.517 & 0.526 \\
    6 & XLM-RoBERTa-large                       & \textbf{0.534} & \textbf{0.551} & \textbf{0.537} & \textbf{0.551} \\ \bottomrule
    \end{tabular}
    \caption{Fine-tuning performance on validation}
    \label{tab:hf_all_results}
\end{table}


Our custom model (Custom BERT-base) for which we used a two-step fine-tuning strategy, once with a language modeling objective to inject domain information and once for task-specific fine-tuning, clearly performed better than multilingual BERT. This shows that even very small amount of domain-specific, code-mixed data improves language model quality, when used for further training. However, it still could not match the performance of XLM-Roberta model which is trained on larger corpora and has a larger Transformer network.

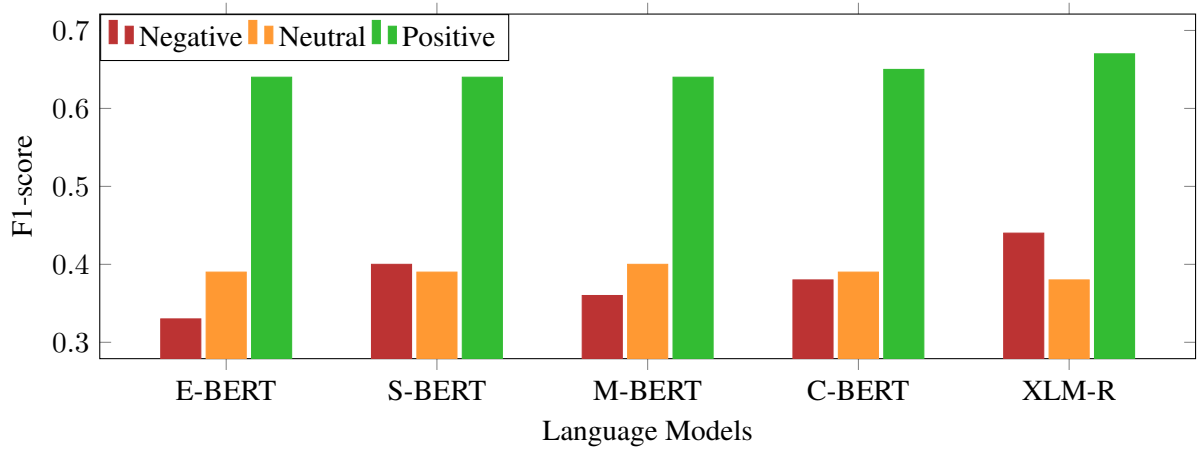
\begin{figure}[h]
\pgfplotsset{width=\textwidth}
\begin{tikzpicture}
    \begin{axis}[
        ybar,
        /pgf/bar width=15pt,
        enlargelimits=0.15,
        legend style={at={(0.187,1)},
          anchor=north,legend columns=-1},
        xlabel={Language Models},
        ylabel={F1-score},
        symbolic x coords={E-BERT, S-BERT, M-BERT, C-BERT, XLM-R},
        xtick=data,
        height=175pt,
        ]
    
        \addplot[red!80!white,fill=red!80!white]
            coordinates {(M-BERT, 0.36)
                         (E-BERT, 0.33)
                         (S-BERT, 0.40)
                         (C-BERT, 0.38)
                         (XLM-R, 0.44)};
        \addplot[orange!80!white,fill=orange!80!white]
            coordinates {(M-BERT, 0.40)
                         (E-BERT, 0.39)
                         (S-BERT, 0.39)
                         (C-BERT, 0.39)
                         (XLM-R, 0.38)};
        \addplot[green!80!white,fill=green!80!white]
            coordinates {(M-BERT, 0.64)
                         (E-BERT, 0.64)
                         (S-BERT, 0.64)
                         (C-BERT, 0.65)
                         (XLM-R, 0.67)};

        \legend{Negative, Neutral, Positive}
    \end{axis}
\end{tikzpicture}
\vspace{-5px}
\caption{F1-scores for each sentiment label on validation data.}
\vspace{-5px}
\label{fig:class}
\end{figure}

Figure~\ref{fig:class} shows the F1-scores, specified per sentiment label. We can see that all models performed similarly on neutral and positive. However, there is much more variation in the negative class. English BERT for example, had particularly poor performance here. XLM-RoBERTa on the other hand, had no issues and even managed to better its score on neutral.

\section{Discussion}
To better understand our results, we performed some error analysis. Since our data consisted of tweets with both English and Spanish words, we expected the multilingual models to perform better than their monolingual counterparts. However, we did not see this pattern in our results. Multilingual BERT model performed better than the English BERT, but worse than the Spanish one. To see in which regard our models differ from each other, we investigated the differences in tokenization, and the effect of language use.



\subsection{Tokenization}
Firstly, we decided to look at the differences between the tokenizers. The pre-trained models in our selection, use either WordPiece (WP) \cite{wu2016google} or SentencePiece (SP) \cite{kudo-richardson-2018-sentencepiece} tokenization, where words are split into substrings (and possibly morphemes). Since Spanish has a more complex morphology system than English \cite{ramirez2010morphological}, the English tokenizer had great difficulty recognizing the correct morphemes. Based on this observation, we hypothesized that fewer wordpieces corresponded to a more accurate tokenization.






\begin{table}[h]
    \centering
    \begin{tabular}{R{0.25cm}lcc}\toprule
      & Tokenizer               & Vocabulary size & Non-first tokens \\\midrule
    1 & English BERT-base       & 30k (WP) & 205,905 \\
    2 & Spanish BERT-base       & 31k (SP)& \textbf{127,323} \\
    3 & Multilingual BERT-base  & 110k (WP) & 152,877 \\
    4 & XLM-RoBERTa large       & 250k (SP) & 131,591 \\\bottomrule
    \end{tabular}
    \caption{Number of non-first tokens per tokenizer on training data.} 
    \label{tab:hf_tokens}
\end{table}

\begin{table}[h]
    \centering
    \small
    \begin{tabular}{R{0.25cm}l}\toprule
      &  Sentence \\\midrule
    1 & 'Since', 'I', 'started', 'working', 'ya', 'ni', 'di', '\#\#s', '\#\#f', '\#\#ru', '\#\#to', 'la', 'v', '\#\#ida', 'lo', '\#\#l' \\
    2 & 'Sin', '\#\#ce', 'I', 'sta', '\#\#r', '\#\#ted', 'w', '\#\#or', '\#\#k', '\#\#ing', 'ya', 'ni', 'disfru', '\#\#to', 'la', 'vida', 'lo', '\#\#l' \\
    3 & 'Since', 'I', 'started', 'working', 'ya', 'ni', 'dis', '\#\#fr', '\#\#uto', 'la', 'vida', 'lo', '\#\#l' \\
    4 & '\_Since', '\_I', '\_started', '\_working', '\_ya', '\_ni', '\_dis', 'fru', 'to', '\_la', '\_vida', '\_lol' \\\bottomrule
    \end{tabular}
    \caption{Wordpieces produced by each model's tokenizer for an example sentence. Each line corresponds to a model: English BERT (1), Spanish BERT (2), Multilingual BERT (3) and XLM-RoBERTa (4).}
    \label{tab:tokenizers}
\end{table}

We therefore came up with an evaluation metric: the number of non-first tokens. Firstly, we tokenized the training data using NLTK word\_tokenize, which is a word-level tokenizer. Secondly, we tokenized the same data using four different BERT tokenizers. We then subtracted the number of tokens for every tokenizer from the number of words, to calculate the amount of non-first tokens. The results of this evaluation can be seen in Table~\ref{tab:hf_tokens}.

Multilingual BERT-base produced considerably fewer tokens than monolingual BERT-base. Again, multilingual XLM-RoBERTa produced fewer non-first tokens than monolingual BERT. These results would indicate that a multilingual tokenizer is better than a monolingual one. However, the monolingual BERT-base Spanish tokenizer breaks this pattern, with the lowest number of non-first tokens.

The increase in number of non-first tokens for the XLM-RoBERTa tokenizer over the multilingual BERT tokenizer might be due to the vocabulary size (i.e., pieces) of the models. XLM-RoBERTa has a vocabulary size of 250k, whereas multilingual BERT uses only a 110k vocabulary. This shows that tokenization has a clear effect on performance although it is not the only determining variable for the overall accuracy, considering the Spanish case. Table \ref{tab:tokenizers} shows an example sentence tokenized by each model's tokenizer. 


\begin{figure}[h]
\pgfplotsset{width=\textwidth}
\begin{tikzpicture}
    \begin{axis}[
        ybar,
        /pgf/bar width=15pt,
        enlargelimits=0.15,
        legend style={at={(0.17,1)},
          anchor=north,legend columns=-1},
        xlabel={Language Models},
        ylabel={Weighted F1},
        symbolic x coords={E-BERT, S-BERT, M-BERT, C-BERT, XLM-R},
        xtick=data,
        height=175pt,
        ymax=0.59
        ]
    
        \addplot[darkblue!80!white,fill=darkblue!80!white]
        	coordinates {(M-BERT, 0.49)
        	             (E-BERT, 0.52)
        	             (S-BERT, 0.50)
        	             (C-BERT, 0.50)
        	             (XLM-R, 0.54)};
        \addplot[darkorange!80!white,fill=darkorange!80!white]
        	coordinates {(M-BERT, 0.43)
        	             (E-BERT, 0.40)
        	             (S-BERT, 0.45)
        	             (C-BERT, 0.45)
        	             (XLM-R, 0.48)};
        \addplot[grey!80!white,fill=grey!80!white]
        	coordinates {(M-BERT, 0.53)
        	             (E-BERT, 0.55)
        	             (S-BERT, 0.57)
        	             (C-BERT, 0.53)
        	             (XLM-R, 0.57)};

        \legend{English, Spanish, Other}
    \end{axis}
\end{tikzpicture}
\vspace{-5px}
\caption{Weighted F1-scores for the Transformer models, separated by language.}
\vspace{-5px}
\label{fig:lang}
\end{figure}
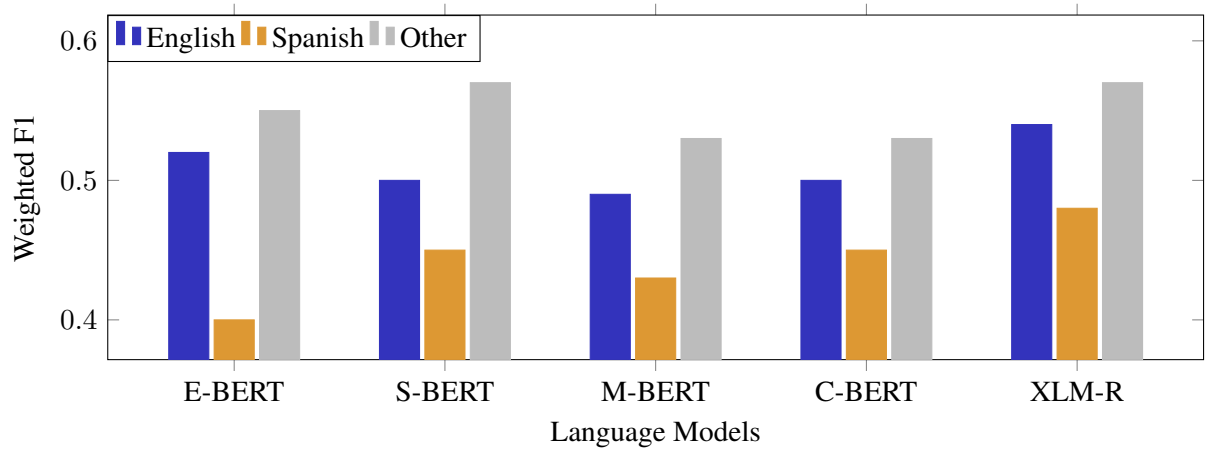

\subsection{Language-specific Performance}
As a second evaluation, we looked at the models' performance on sentences with a different ratio of languages (English and Spanish). For this, we split the development dataset into sentences with more Spanish than English words, and vice-versa by using token-level language labels. We also present another group, with miscellaneous labels such as ambiguous and other \footnote{We used a threshold of 0.75 for both the English and Spanish splits, and 0.4 for `Others'.}. We then looked at the predictions, and calculated the weighted F1-score for every language group. The results are shown in Figure~\ref{fig:lang}. 

As one would expect, when looking at the monolingual Transformers, we see that the Spanish and English BERT models excel at their respective language group. However, while English BERT suffered from very poor performance on predominantly Spanish tweets, its Spanish counterpart had a more balanced performance. Interestingly, although multilingual BERT's performance is on-par with Spanish BERT on the English group, it underperforms on the Spanish group, which would explain why Spanish BERT has a better overall F1-score than its multilingual counterpart.


For the custom model (C-BERT), a two-step fine-tuning strategy to enrich the model with code-switched domain-specific information (i.e., social media) improved performance over the multilingual BERT. As shown in Figure 3, C-BERT performed better on especially Spanish group of sentences, compared to the it's multilingual base.
This indicates that more domain-specific training could increase the quality of a multilingual pre-trained model, considering the task and code-mixing challenge.

Finally, Figure~\ref{fig:lang} clearly shows why XLM-RoBERTa outperforms all of the other models. It has the best performance on all three groups of sentences, regardless of the language. 

\section{Conclusion}

In this paper we presented FiSSA, our approach for sentiment classification on Spanish-English data. We showed that fine-tuning a pre-trained language model is a good alternative to a standard model, especially when the amount of labeled training data is limited. By fine-tuning XLM-RoBERTa, we achieved a weighted F1-score of 0.537 on development data and 0.739 on test data

In the discussion, we evaluated the effect of tokenization and language-specific performance of each model to better understand the overall results. 
\clearpage

\bibliographystyle{acl}
\bibliography{coling2020}

\begin{thebibliography}{}

\bibitem[\protect\citename{Akbik \bgroup et al.\egroup }2019]{akbik2019naacl}
Alan Akbik, Tanja Bergmann, and Roland Vollgraf.
\newblock 2019.
\newblock Pooled contextualized embeddings for named entity recognition.
\newblock In {\em {NAACL} 2019, 2019 Annual Conference of the North American
  Chapter of the Association for Computational Linguistics}, page 724–728.

\bibitem[\protect\citename{Bar and Dershowitz}2014]{bar-dershowitz-2014-tel}
Kfir Bar and Nachum Dershowitz.
\newblock 2014.
\newblock The {T}el aviv university system for the code-switching workshop
  shared task.
\newblock In {\em Proceedings of the First Workshop on Computational Approaches
  to Code Switching}, pages 139--143, Doha, Qatar, October. Association for
  Computational Linguistics.

\bibitem[\protect\citename{Cañete \bgroup et al.\egroup }2020]{CaneteCFP2020}
José Cañete, Gabriel Chaperon, Rodrigo Fuentes, and Jorge Pérez.
\newblock 2020.
\newblock Spanish pre-trained bert model and evaluation data.
\newblock In {\em to appear in PML4DC at ICLR 2020}.

\bibitem[\protect\citename{Conneau \bgroup et al.\egroup
  }2019]{conneau2019unsupervised}
Alexis Conneau, Kartikay Khandelwal, Naman Goyal, Vishrav Chaudhary, Guillaume
  Wenzek, Francisco Guzmán, Edouard Grave, Myle Ott, Luke Zettlemoyer, and
  Veselin Stoyanov.
\newblock 2019.
\newblock Unsupervised cross-lingual representation learning at scale.

\bibitem[\protect\citename{Devlin \bgroup et al.\egroup }2018]{devlin2018bert}
Jacob Devlin, Ming{-}Wei Chang, Kenton Lee, and Kristina Toutanova.
\newblock 2018.
\newblock {BERT:} pre-training of deep bidirectional transformers for language
  understanding.
\newblock {\em CoRR}, abs/1810.04805.

\bibitem[\protect\citename{Hochreiter and
  Schmidhuber}1997]{doi:10.1162/neco.1997.9.8.1735}
Sepp Hochreiter and Jürgen Schmidhuber.
\newblock 1997.
\newblock Long short-term memory.
\newblock {\em Neural Computation}, 9(8):1735--1780.

\bibitem[\protect\citename{Kudo and
  Richardson}2018]{kudo-richardson-2018-sentencepiece}
Taku Kudo and John Richardson.
\newblock 2018.
\newblock {S}entence{P}iece: A simple and language independent subword
  tokenizer and detokenizer for neural text processing.
\newblock In {\em Proceedings of the 2018 Conference on Empirical Methods in
  Natural Language Processing: System Demonstrations}, pages 66--71, Brussels,
  Belgium, November. Association for Computational Linguistics.

\bibitem[\protect\citename{Lample and Conneau}2019]{lample2019cross}
Guillaume Lample and Alexis Conneau.
\newblock 2019.
\newblock Cross-lingual language model pretraining.
\newblock {\em Advances in Neural Information Processing Systems (NeurIPS)}.

\bibitem[\protect\citename{Liu \bgroup et al.\egroup }2019]{liu2019roberta}
Yinhan Liu, Myle Ott, Naman Goyal, Jingfei Du, Mandar Joshi, Danqi Chen, Omer
  Levy, Mike Lewis, Luke Zettlemoyer, and Veselin Stoyanov.
\newblock 2019.
\newblock Roberta: A robustly optimized bert pretraining approach.
\newblock {\em arXiv preprint arXiv:1907.11692}.

\bibitem[\protect\citename{Patwa \bgroup et al.\egroup
  }2020]{patwa2020sentimix}
Parth Patwa, Gustavo Aguilar, Sudipta Kar, Suraj Pandey, Srinivas PYKL,
  Bj{\"o}rn Gamb{\"a}ck, Tanmoy Chakraborty, Thamar Solorio, and Amitava Das.
\newblock 2020.
\newblock Semeval-2020 task 9: Overview of sentiment analysis of code-mixed
  tweets.
\newblock In {\em Proceedings of the 14th International Workshop on Semantic
  Evaluation ({S}em{E}val-2020)}, Barcelona, Spain, December. Association for
  Computational Linguistics.

\bibitem[\protect\citename{Pires \bgroup et al.\egroup
  }2019]{pires-etal-2019-multilingual}
Telmo Pires, Eva Schlinger, and Dan Garrette.
\newblock 2019.
\newblock How multilingual is multilingual {BERT}?
\newblock In {\em Proceedings of the 57th Annual Meeting of the Association for
  Computational Linguistics}, pages 4996--5001, Florence, Italy, July.
  Association for Computational Linguistics.

\bibitem[\protect\citename{Ramirez \bgroup et al.\egroup
  }2010]{ramirez2010morphological}
Gloria Ramirez, Xi~Chen, Esther Geva, and Heidi Kiefer.
\newblock 2010.
\newblock Morphological awareness in spanish-speaking english language
  learners: Within and cross-language effects on word reading.
\newblock {\em Reading and Writing}, 23(3-4):337--358.

\bibitem[\protect\citename{Solorio and Liu}2008]{solorio2008}
Thamar Solorio and Yang Liu.
\newblock 2008.
\newblock Part-of-speech tagging for english-spanish code-switched text.
\newblock pages 1051--1060, 01.

\bibitem[\protect\citename{Solorio \bgroup et al.\egroup
  }2014]{solorio-etal-2014-overview}
Thamar Solorio, Elizabeth Blair, Suraj Maharjan, Steven Bethard, Mona Diab,
  Mahmoud Ghoneim, Abdelati Hawwari, Fahad AlGhamdi, Julia Hirschberg, Alison
  Chang, and Pascale Fung.
\newblock 2014.
\newblock Overview for the first shared task on language identification in
  code-switched data.
\newblock In {\em Proceedings of the First Workshop on Computational Approaches
  to Code Switching}, pages 62--72, Doha, Qatar, October. Association for
  Computational Linguistics.

\bibitem[\protect\citename{Vaswani \bgroup et al.\egroup
  }2017]{DBLP:journals/corr/VaswaniSPUJGKP17}
Ashish Vaswani, Noam Shazeer, Niki Parmar, Jakob Uszkoreit, Llion Jones,
  Aidan~N. Gomez, Lukasz Kaiser, and Illia Polosukhin.
\newblock 2017.
\newblock Attention is all you need.
\newblock {\em CoRR}, abs/1706.03762.

\bibitem[\protect\citename{Vilares \bgroup et al.\egroup
  }2015]{vilares-etal-2015-sentiment}
David Vilares, Miguel~A. Alonso, and Carlos G{\'o}mez-Rodr{\'\i}guez.
\newblock 2015.
\newblock Sentiment analysis on monolingual, multilingual and code-switching
  twitter corpora.
\newblock In {\em Proceedings of the 6th Workshop on Computational Approaches
  to Subjectivity, Sentiment and Social Media Analysis}, pages 2--8, Lisboa,
  Portugal, September. Association for Computational Linguistics.

\bibitem[\protect\citename{Vilares \bgroup et al.\egroup
  }2016]{vilares-etal-2016-en}
David Vilares, Miguel~A. Alonso, and Carlos G{\'o}mez-Rodr{\'\i}guez.
\newblock 2016.
\newblock {EN}-{ES}-{CS}: An {E}nglish-{S}panish code-switching twitter corpus
  for multilingual sentiment analysis.
\newblock In {\em Proceedings of the Tenth International Conference on Language
  Resources and Evaluation ({LREC}'16)}, pages 4149--4153, Portoro{\v{z}},
  Slovenia, May. European Language Resources Association (ELRA).

\bibitem[\protect\citename{Wolf \bgroup et al.\egroup
  }2019]{Wolf2019HuggingFacesTS}
Thomas Wolf, Lysandre Debut, Victor Sanh, Julien Chaumond, Clement Delangue,
  Anthony Moi, Pierric Cistac, Tim Rault, R'emi Louf, Morgan Funtowicz, and
  Jamie Brew.
\newblock 2019.
\newblock Huggingface's transformers: State-of-the-art natural language
  processing.
\newblock {\em ArXiv}, abs/1910.03771.

\bibitem[\protect\citename{Wu \bgroup et al.\egroup }2016]{wu2016google}
Yonghui Wu, Mike Schuster, Zhifeng Chen, Quoc~V Le, Mohammad Norouzi, Wolfgang
  Macherey, Maxim Krikun, Yuan Cao, Qin Gao, Klaus Macherey, et~al.
\newblock 2016.
\newblock Google's neural machine translation system: Bridging the gap between
  human and machine translation.
\newblock {\em arXiv preprint arXiv:1609.08144}.

\bibitem[\protect\citename{Yang \bgroup et al.\egroup }2019]{yang2019xlnet}
Zhilin Yang, Zihang Dai, Yiming Yang, Jaime Carbonell, Ruslan Salakhutdinov,
  and Quoc~V. Le.
\newblock 2019.
\newblock Xlnet: Generalized autoregressive pretraining for language
  understanding.
\newblock cite arxiv:1906.08237Comment: Pretrained models and code are
  available at https://github.com/zihangdai/xlnet.

\end{thebibliography}

\end{document}